\documentclass[runningheads]{llncs}
\usepackage[T1]{fontenc}
\usepackage{cite}
\usepackage{hyperref}
\usepackage{amsmath,amssymb,amsfonts,nccmath}
\usepackage{bbm}
\usepackage{algpseudocode}
\usepackage{adjustbox}
\usepackage[linesnumbered,ruled,vlined]{algorithm2e}
\usepackage{multirow}
\usepackage[table,xcdraw]{xcolor}
\usepackage{textcomp}
\usepackage{xcolor}
\usepackage{graphicx}
\usepackage[bb=px]{mathalfa} 
\usepackage{tikz}
\usetikzlibrary{positioning, fit, arrows.meta, shapes}
\tikzstyle{line}=[draw]
\usepackage{enumerate}
\usepackage{longtable}
\usepackage{latexsym}
\usepackage{url}
\usepackage[switch]{lineno}
\usepackage[inline]{enumitem}
\usepackage{lipsum}
\usepackage{footnote}
\usepackage{caption}
\usepackage{subcaption}
\usepackage{rotating}
\usepackage{tabto}
\usepackage{comment}
\begin{document}
\title{Data Augmentation in Graph Neural Networks: \\The Role of Generated Synthetic Graphs}

\author{Sümeyye Baş\inst{1,3} \and
Kıymet Kaya\inst{2,3,4} \and
Resul Tugay \inst{5}\and \\
Şule Gündüz Öğüdücü \inst{1,3}}

\institute{
Istanbul Technical University, Department of Artificial Intelligence and Data Engineering, Istanbul, Turkey \and
ITU, Department of Computer Engineering, Istanbul, Turkey \and
ITU, AI Research and Application Center, Istanbul, Turkey \and
BTS Group, Istanbul, Turkey \and
Gazi University, Department of Computer Engineering, Ankara, Turkey \\
\email{bass20@itu.edu.tr, kayak16@itu.edu.tr, resultugay@gazi.edu.tr, sgunduz@itu.edu.tr} \\
}

\maketitle
\begin{abstract} 

Graphs are crucial for representing interrelated data and aiding predictive modeling by capturing complex relationships. Achieving high-quality graph representation is important for identifying linked patterns, leading to improvements in Graph Neural Networks (GNNs) to better capture data structures. However, challenges such as data scarcity, high collection costs, and ethical concerns limit progress. As a result, generative models and data augmentation have become more and more popular. This study explores using generated graphs for data augmentation, comparing the performance of combining generated graphs with real graphs, and examining the effect of different quantities of generated graphs on graph classification tasks. The experiments show that balancing scalability and quality requires different generators based on graph size. Our results introduce a new approach to graph data augmentation, ensuring consistent labels and enhancing classification performance.
\end{abstract}

\section{Introduction}

Graphs serve as important representations of interrelated data in various fields, including social networks and chemical sciences, due to their ability to encapsulate complex relationships and facilitate critical tasks such as predictive modeling. Obtaining high-quality graph representations is vital for revealing underlying interrelated patterns and phenomena in fields that rely on graph-related data \cite{Tan2019}. 

However, the progress in this area is being hampered by the lack of useful big datasets and consistent evaluation processes. Despite significant progress in data availability in recent years, this development remains limited in various application domains for privacy and security reasons \cite{tedersoo2021data, lu2024machine}. This indicates that future models may intentionally or unintentionally resort to generated synthetic data. 

\begin{figure*}[ht]
\centering
\includegraphics[width=\linewidth]{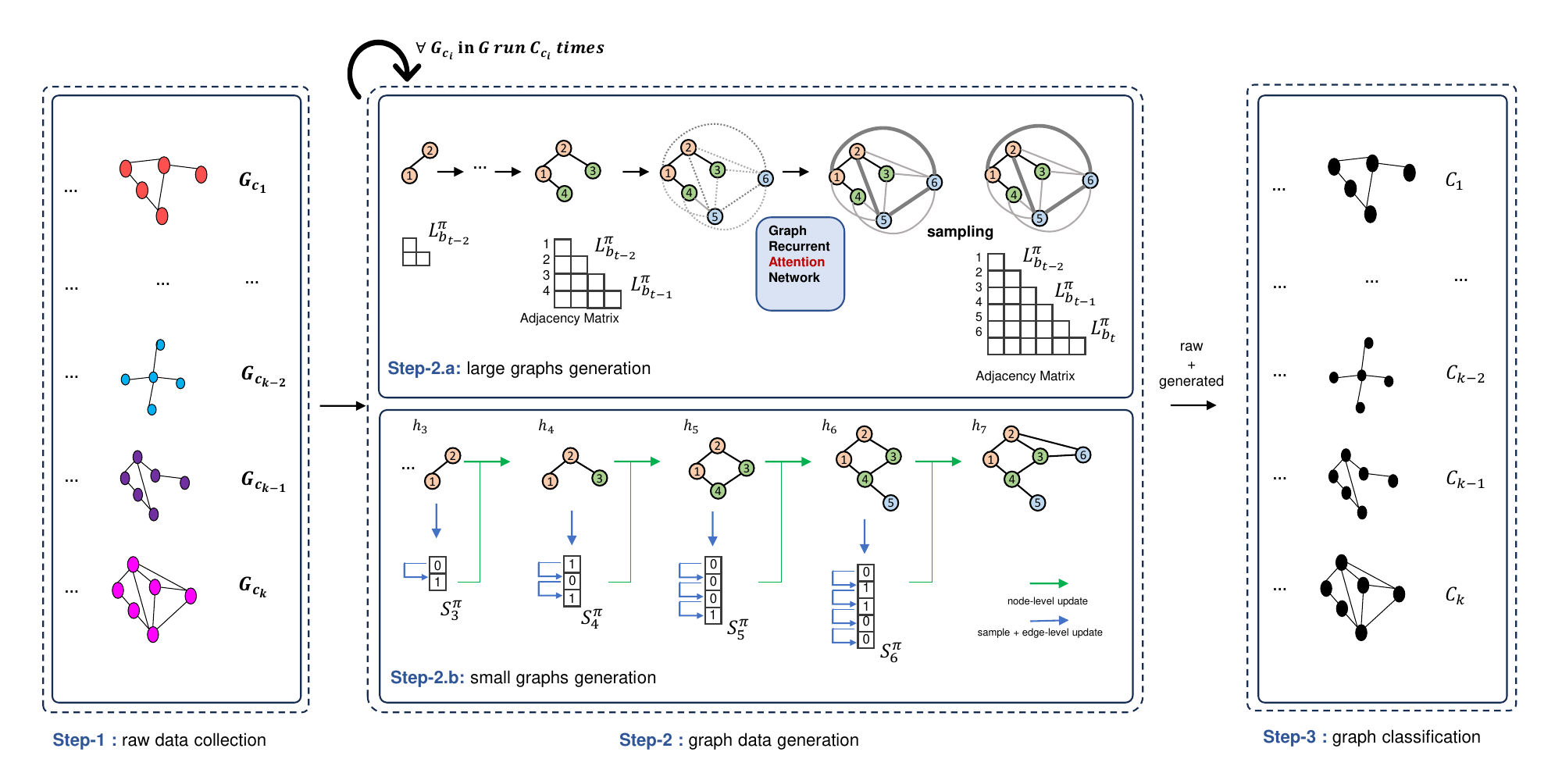}
\caption{Graph Classification with Graph Size-Aware Data Augmentation.}

\label{fig_proposed}
\end{figure*}

The generation of synthetic data is becoming more and more popular for a variety of reasons such as improving data diversity, and enhancing privacy and security \cite{villalobos2022run}. In many domains, it is often difficult to gather sufficient data tailored to specific tasks. Data augmentation emerges as a crucial solution to this challenge. In research areas, involving graph-related data, leveraging proper synthetic graphs proves to be more efficient, cost-effective, and straightforward compared to acquiring additional real-world datasets. Moreover, it offers enhanced privacy protections, particularly in sensitive industries such as healthcare and finance \cite{alemohammad2023selfconsuming}.

The \textit{Graph Classification with Graph Size-Aware Data Augmentation} framework we propose in this study is presented in Fig. \ref{fig_proposed}. It's important to clarify that our approach involves data augmentation, encompassing both synthetic and real-world data, such that, the amount of the training set is increased, rather than relying solely on synthetic data. The proposed approach emerges from extensive literature research, revealing the diverse capabilities of different graph generators \cite{touat2023gran,you2018graphrnn}. Recognizing that raw datasets possess varying characteristics, which can influence generator performance, we examined these generators especially for graph size sensitivity which led us to introduce a novel framework aimed at enhancing graph classification. The framework operates by splitting datasets according to graph labels, training individual generator models for each class, and leveraging both original and generated graphs in the graph classification process. Experimental results demonstrate that this tailored approach significantly boosts classification performance, particularly for datasets with a limited number of graphs.

The main contributions of this study can be summarized as follows:
\begin{itemize}

    \item Examining the usability of generated synthetic data to improve graph classification model performances. 

    \item Investigating how the ratio of generated data affects the prediction capabilities of the models by training the models with different proportions of synthetic data.

    \item Addressing graph labeling issue in graph data augmentation.

    \item Questioning whether it is worth spending time and resources to collect additional real data, comparing the improvements provided by real-world and synthetic data for data augmentation.

\end{itemize}

The rest of the paper is organized as follows. Section \ref{sec:2} presents related works and Section \ref{sec:3} gives details of the methodology. Section \ref{sec:4} presents the experimental results of the study. Lastly, Section \ref{sec:6} concludes the paper.

\section{Literature Review} \label{sec:2}

Data augmentation has recently drawn significant attention in the field of machine learning, especially for deep learning models,\cite{ding2022data,zhao2023graph,lin2023spectral}, thanks to its ability to enhance model performance and generalization by incorporating additional training data. Data augmentation is commonly used in computer vision and text applications. Data augmentation with images is relatively simpler due to the Euclidean nature and well-organized structure of image instances. Pixel matrices can be easily transformed using common rule-based techniques like rotation, scaling, and flipping preserving the labels \cite{bansal2023leaving}. Similarly, for text-level data augmentation rule-based data augmentation approaches including random insertion, random deletion, and synonym substitution are quite effective \cite{liu2020survey}. AugGPT rephrases sentences in the training data through rule-based approaches to increase the training data size and ensure accurate labeling in the generated text data \cite{dai2023auggpt}. 

However, compared to text and image data types graphs are irregular and non-Euclidean. Therefore, even small changes in the structure may result in a loss of inter-related information, making the augmentation process more complex for graph data. Graph Data Augmentation methods can be broadly examined under two headings: rule-based and learned (AI-based) approaches. Among these, rule-based methods apply predefined rules to modify graph data, such as random edge removal and graph clipping while learned methods, such as graph structure learning and graph rationalization, exploit learnable parameters for data augmentation that can be trained independently or with downstream tasks. 

The conventional rule-based graph augmentation techniques widely applied in the literature, such as arbitrary node removal, edge modification, or the occlusion of node characteristics, rely on random alterations of network structures or attributes. However, these arbitrary changes often compromise label invariance by inadvertently damaging significant label-related information, thus failing to generate appropriate graph data and improve graph prediction models' performance in practical applications. To address these limitations, GraphAug is offered as a solution by computing label-invariant augmentations through an automated approach, thereby safeguarding crucial label-related data within graph datasets \cite{luo2023automated}. Furthermore, Sui et al. proposed Adversarial Invariant Augmentation (AIA), a strategy aimed at mitigating covariate shifts in graph representation learning \cite{sui2023unleashing}. Yue et al. advanced a label-invariant augmentation method for graph-structured data in graph contrastive learning, generating augmented samples in challenging directions within the representation space while maintaining the original sample labels \cite{yue2022labelinvariant}. 

Despite advances in data augmentation, studies on using generated data for prediction in graph neural networks are limited in the literature \cite{he2023synthetic}. Zero-shot image classification was performed using data generated with models that had been extensively trained. The performance of prediction models using synthetic data and real data for image classification is compared and similar accuracy values are observed \cite{besnier2019dataset}. The effect of synthetic graphs on the model performance of node classification algorithms was examined, and small improvements in performance were obtained through pre-training using graphs with similar characteristics. Sun et al. presented the MoCL framework for learning molecular representations, utilizing both local and global domain expertise to guide the augmentation procedure and guarantee variation without changing graph semantics, as shown on several molecular datasets \cite{sun2022mocl}. Using social network graph datasets, Tang et al. performed cosine similarity-based cross-operation on the initial characteristics to produce additional graph features for node classification tasks \cite{tang2021data}. In this study, we investigate the impact of graph data augmentation on the graph classification task. Generator methods proposed in the literature for data augmentation offer different advantages. Our work differs from its contemporaries in that it appropriately combines two state-of-the-art generator models, according to their advantages in terms of graph size.

\section{Methodology} \label{sec:3}

The overall structure of our proposed method is depicted in Fig. \ref{fig_proposed}. First, for the graph classification task, the data $G = {G_1,..., G_s}$ with $E$ edges and $V$ nodes, is grouped as small or large according to the average number of nodes and edges of the graphs it contains. In this study, the average number of nodes is fifty and above, or an edge number of thousand two hundred twenty-five (edge numbers for a fully connected simple graph with fifty nodes), and above is determined as large, otherwise small. In Step 2 of Fig. \ref{fig_proposed}, the appropriate generation model is selected according to whether the graphs in the data are small or large, and for each graph class $C_i$, the desired number of synthetic graphs for that class $C_{C_i}$ is generated for data augmentation. Here we recommend GRAN for data consisting of large graphs and GraphRNN for small ones. GRAN, with its focus on balancing efficiency and quality—particularly through its stride parameter—may experience a reduction in generation quality for smaller graphs when compared to GraphRNN. On the other hand, GRAN's methodology is optimized for handling larger graphs, whereas GraphRNN's architecture and training process are better suited for generating smaller-scale graphs \cite{touat2023gran} and mostly give out of memory error for larger graphs despite efforts to enhance scalability through techniques like bread-first-search (BFS) node ordering scheme. Graph generation methods GraphRNN, GRAN, and all the classification methods used during the experiments are detailed in the subheadings, and the code repository is also available here\footnote{https://github.com/sumeyyebas/AIGraphAugmentation}.

\subsection{Graph Data Generation}

Learning a distribution $p_{model}(G)$ over graphs is the aim of generative model learning. This is achieved by sampling a collection of observed graphs $G = {G_1,..., G_s}$ from the data distribution $p(G)$, where each graph in $G$ may vary in the number of nodes and edges \cite{you2018graphrnn}. Instead of directly acquiring knowledge about the probability distribution $p(G)$, which is difficult to define precisely the representation of the sample space, an auxiliary random variable $\pi$ is sampled to represent node ordering as sequences. This transforms the graph generation process into the generation of node and edge sequences, where nodes and edges are generated autoregressively. An adjacency matrix with a node ordering $\pi$ maps nodes to rows and columns of the matrix enabling each graph in $G$

to be depicted by the adjacency matrix $A^\pi \in \mathbb{R}^{n \times n}$ \ref{adjmatrix_inpiorder}. 

\begin{equation}
A_{i, j}^\pi=\mathbbm{1}\left[\left(\pi\left(v_i\right), \pi\left(v_j\right)\right) \in E\right].
\label{adjmatrix_inpiorder}
\end{equation}

The aforementioned generator models were designed to work with simple graphs $G = (V, E)$. Initially, graph nodes and edges are represented as sequences and sequences of sequences using a mapping $f_S$,  respectively. For a graph $ G $ sampled from $ p(G) $ with $ n $ nodes under a node ordering $ \pi $, the sequence $ S_\pi $ is obtained as in Equation \ref{functiongunderpi}, where $S_{i}^\pi$ is an adjacency vector representing the edges between node $ \pi(v_i) $ and the preceding nodes $ \pi(v_j) $, $ j \in \{1, ..., i-1\} $, already present in the graph \ref{Si_by_adjmatrix}. 

\begin{equation}
f_S(G, \pi) = (S_1^\pi, ..., S_n^\pi) 
\label{functiongunderpi}
\end{equation}

\begin{equation}
S_i^\pi=\left(A_{1, i}^\pi, \ldots, A_{i-1, i}^\pi\right)^T, \forall i \in\{2, \ldots, n\}
\label{Si_by_adjmatrix}
\end{equation}

\begin{equation}
\resizebox{0.45\hsize}{!}{
$p(G)=\sum\limits_{S^{\pi}} p\left(S^\pi\right) \mathbbm{1}\left[f_G\left(S^\pi\right)=G\right]$
}
\label{final_pg}
\end{equation}

In the case of undirected graphs, $S^{\pi}$ uniquely determines a graph $G$, denoted by the mapping $f_G(\cdot)$, where  $f_{G}(S^{\pi}) = G$. This sequentialization process allows generators to observe $S^\pi$ and learn about its probability distribution, $p(S^\pi)$, which can be analyzed sequentially as $S^{\pi}$ exhibits a sequential nature. During inference time, generators can derive samples of $G$ without explicitly calculating $p(G)$ by sampling $S^\pi$, which corresponds to $G$ through the function $f_G$. With these concepts, $p(G)$ can be expressed as a marginal probability distribution of the joint distribution $p(G, S^\pi)$ in Equation \ref{final_pg}, where $p(S^\pi)$ is the distribution that the generator aims to learn.

\subsubsection{Large Graphs Generation: GRAN \cite{liao2020efficient}} \label{subsec:largegraphgen} 

The overall procedure of a generation phase with GRAN is illustrated in Step-2.a of Fig. \ref{fig_proposed}. GRAN promises to provide a strong autoregressive conditioning between the graph's generated and to-be-generated portions as attention-based GNN helps better distinguish multiple newly added nodes. While expressing networks as adjacency matrices, some matrices remain unchanged under certain permutations, resulting in symmetry. To solve this, GRAN constructs a set of symmetric permutations as in Equation \ref{gran_0} and develops a surjective function $u$ that maps permutations to symmetric permutations. Thus, for a graph $G$, different adjacency matrices for all permutations are modeled. However, for undirected graphs, it is enough to just model the lower triangular portion of the adjacency matrix $L^\pi$. GRAN generates the lower triangular component $L^\pi$ block by block, adding one block of nodes and associated edges at a time $t$. This procedure considerably decreases auto-regressive graph creation decisions by a factor of $ O(N) $, where $ N = |V|$. 

\begin{equation}
\resizebox{0.3\hsize}{!}{
$
\Delta(A^{\pi}) = \left\{ \tilde{\pi} \mid A^{\tilde{\pi}} = A^{\pi} \right\}
$
}
\label{gran_0}
\end{equation}

\begin{equation}
\resizebox{0.35\hsize}{!}{
$b_{t} = \{B(t - 1) + 1,..., Bt\}$.
}
\label{gran_1}
\end{equation}

\begin{equation}
\resizebox{0.45\hsize}{!}{
 $p(L^{\pi}) = \Pi_{t=1}^{T} p(L^{\pi}_{b_t} | L^{\pi}_{b_1}, \cdot, L^{\pi}_{b_{t-1}})$ 
}
\label{gran_2}
\end{equation}

GRAN creates one block of B rows of $ L^{\pi} $ at a time. The $t$-th block consists of rows with indices as in Equation \ref{gran_1}. The number of steps required to create a graph is therefore $T = O(N/B)$. The conditional probability in Equation \ref{gran_2} determines the likelihood of producing the current block. The probability function it needs to learn becomes a long conditional probability as it uses previous blocks to infer the next block. To avoid long-term bottlenecks and use the structural features of graphs, GRAN prefers GNNs over RNNs.

\subsubsection{Small Graphs Generation: GraphRNN\cite{you2018graphrnn}} \label{subsec:smallgraphgen} 

In GraphRNN, the graph sequentialization is followed by constructing a scalable auto-regressive model that is suitable for small-medium size of graphs and can benefit from graph structure. Its generation process is shown in the Step-2.b of Fig. \ref{fig_proposed}. GraphRNN can be viewed as a hierarchical model where new nodes are constructed by a graph-level RNN and the edges of each newly formed node are generated by an edge-level RNN, all while maintaining the state of the graph.

\begin{equation}
\resizebox{0.32\hsize}{!}{
$h_i = f_{\text{trans}}(h_{i-1}, S_{i-1}^\pi)$}
\label{graphrnn_h}
\end{equation}

\begin{equation}
\resizebox{0.18\hsize}{!}{
$\theta_i = f_{\text{out}}(h_i)$}
\label{graphrnn_theta}
\end{equation}

\begin{equation}
\resizebox{0.32\hsize}{!}{
$S^\pi = f_S(G, BFS(G,\pi ))$}
\label{eq:23}
\end{equation}

The probability distribution $ p(S_i\pi | S_{<i}\pi) $ for each $ i$ is intricate, requiring an understanding of how node $ \pi(v_i) $ connects to preceding nodes based on the previously added nodes. GraphRNN suggests parameterizing $ p(S_i\pi | S_{<i}\pi) $ with a neural networks model ensuring scalability with share weights across all time steps. GraphRNN employs an RNN comprising a state-transition function $f_{\text{trans}}$  and an output function $f_{\text{out}}$ as in Equations \ref{graphrnn_h} and \ref{graphrnn_theta}, where $h_i$ in $ \mathbb{R}^d $ represents the state encoding of the generated graph up to this point, $ S_{i-1} $ is the adjacency vector for the most recently generated node $i-1$, and $i$ denotes the distribution of the adjacency vector for the next node (i.e., $S_i$ follows distribution $ P_o $). Generally, $ f_{\text{trans}} $ and $ f_{\text{out}} $ can be any neural network, and $ P_o $ can be any distribution over binary vectors. 

A key finding of the GraphRNN technique is that, without sacrificing generality, it learns to produce graphs using breadth-first-search (BFS) node orderings instead of learning to generate graphs under any conceivable node permutation. BFS also provide a unique representation of graphs. Therewith, Equation \ref{functiongunderpi} is changed to Equation \ref{eq:23}, with the deterministic BFS function represented by $BFS(\cdot)$. Specifically, this BFS function takes a random permutation $i$ as its input, selects node $v1$ as the starting point, and appends each node's neighbors to the BFS queue following the order given by the permutation. It should be noted that the BFS function is many-to-one, meaning that multiple permutations can result in the same ordering once the BFS function is executed.

\subsection{Graph Classification} \label{subsec:graphclassification}

\noindent \indent
\textbf{GraphSAGE} (Graph Sample and Aggregation) \cite{hamilton2018inductive} creates node embeddings by sampling and aggregating data from each node's neighborhood. GraphSAGE generates robust graph categorization representations by combining information from several layers of the graph.

\textbf{GIN0:} GIN (Graph Isomorphism Network) generates node embeddings by using message-passing procedures and a learnable set function through a multi-layer perceptron network. GIN is suitable for graph classification applications since it can capture higher-order graph structures. GIN0 sets the learnable $\varepsilon$ parameter as 0 and depends rather on a set aggregation algorithm for updating node features. Therefore, it is computationally cheaper but less flexible \cite{xu2019powerful}.

\textbf{GINWithJK} \cite{xu2019powerful} adds the concept of jumping knowledge (JK) concept to the GIN model. This concept combines representations from several layers to improve the final node embeddings. Therefore, better information flow between layers can be achieved.

\textbf{GCNWithJK} \cite{kipf2017semisupervised} Graph Convolutional Networks (GCN) iteratively aggregate information from neighboring nodes. GCN with Jumping Knowledge (GCNWithJK) directly merges node representations from several GCN layers. So, capturing both local and global graph structures becomes easier. 

\textbf{EdgePool} \cite{diehl2019edge} is an edge-level graph pooling technique utilized in Graph Neural Networks (GNNs). It efficiently reduces the size of the graph while maintaining important structural information by selectively aggregating edges. Along with this, every EdgePool layer outputs the mapping between every node in the old graph and every node in the newly-pooled graph. An inverse mapping from pooled nodes to unpooled nodes is produced during unpooling. It is possible to link this mapping via many pooling layers because each node is assigned to exactly one merged node. 

\section{Experimental Results}  \label{sec:4}

We observed the effect of synthetic graph data by GNNs on graph classification on six public datasets from TU\footnote{https://chrsmrrs.github.io/datasets/docs/datasets/} - three chemical compounds (MUTAGENICITY, ENZYMES, MUTAG), two social networks (COLLAB, TWITCH EGOS), and one protein interactions (DD). The descriptive statistics of these benchmark datasets are given in Table \ref{tab:data}.

\begin{table*}[ht]
\caption{Benchmark Graph Datasets Statistics}
\renewcommand{\arraystretch}{1.1}
\label{tab:data}
\centering
\resizebox{.95\linewidth}{!}{
\begin{tabular}{llccccccc}
\hline
name & \#graphs & \begin{tabular}[c]{@{}c@{}}avg. \\ nodes\end{tabular} & \begin{tabular}[c]{@{}c@{}}avg. \\ edges\end{tabular} & \begin{tabular}[c]{@{}c@{}}avg. \\ degree\end{tabular} & \begin{tabular}[c]{@{}c@{}}avg. \\ density\end{tabular} & \begin{tabular}[c]{@{}c@{}}avg. \\ diameter\end{tabular} & \#classes & \begin{tabular}[c]{@{}c@{}}class \\ distributions \\ (\%)\end{tabular} \\ \hline
DD & 518 & 258.74 & 650.23 & 4.99 & 0.0232 & 19.75 & 2 & 63-36 \\ \hline
COLLAB & 4001 & 73.40 & 2357.18 & 36.97 & 0.5076 & 1.87 & 3 & 51-32-15 \\ \hline
TWITCH EGOS & 101894 & 29.69 & 86.51 & 5.39 & 0.2020 & 2.00 & 2 & 53-46 \\ \hline
MUTAGENICITY & 3467 & 29.52 & 30.51 & 2.05 & 0.0921 & 9.92 & 2 & 55-44 \\ \hline
ENZYMES & 360 & 32.28 & 61.04 & 3.83 & 0.1592 & 11.30 & 6 & 16-16-16-16-16-16 \\ \hline
MUTAG & 152 & 17.95 & 19.79 & 2.19 & 0.1383 & 8.24 & 2 & 66-33 \\ \hline

\end{tabular}
}
\end{table*}

\begin{table*}[ht!]
\caption{Graph Classification Results - I}
\renewcommand{\arraystretch}{1.2}
\label{tab:res}
\centering
\resizebox{.99\linewidth}{!}{
\begin{tabular}{clcccccccccc}
\hline
\textbf{} &  & \multicolumn{2}{c}{\textbf{GraphSAGE}} & \multicolumn{2}{c}{\textbf{GIN0}} & \multicolumn{2}{c}{\textbf{GINWithJK}} & \multicolumn{2}{c}{\textbf{GCNWithJK}} & \multicolumn{2}{c}{\textbf{EdgePool}} \\ \hline
\textbf{} &  & \textbf{Acc.} & \textbf{Epoch} & \textbf{Acc.} & \textbf{Epoch} & \textbf{Acc.} & \textbf{Epoch} & \textbf{Acc.} & \textbf{Epoch} & \textbf{Acc.} & \textbf{Epoch} \\ \hline
 & raw-data & 0.630 & 6 & 0.605 & 6 & 0.545 & 27 & 0.630 & 5 & 0.660 & 4 \\ \cline{2-12} 
 & w/ Real =  $ |R|$ & 0.630 & 12 & \textbf{0.665} & 40 & 0.630 & 12 & 0.630 & 4 & 0.630 & 5 \\ \cline{2-12} 
 & \cellcolor[HTML]{F3F2C5}$\textbf{w/ Gen.}_{1}$ = $|R|$ & \cellcolor[HTML]{F3F2C5}0.635 & \cellcolor[HTML]{F3F2C5}10 & \cellcolor[HTML]{F3F2C5}0.650 & \cellcolor[HTML]{F3F2C5}14 & \cellcolor[HTML]{F3F2C5}0.665 & \cellcolor[HTML]{F3F2C5}20 & \cellcolor[HTML]{F3F2C5}0.630 & \cellcolor[HTML]{F3F2C5}4 & \cellcolor[HTML]{F3F2C5}0.630 & \cellcolor[HTML]{F3F2C5}5 \\ \cline{2-12} 
 & \cellcolor[HTML]{F3F2C5}$\textbf{w/ Gen.}_{2}$ = $|R|*2$ & \cellcolor[HTML]{D3D050}
\textbf{0.695} & \cellcolor[HTML]{F3F2C5}31 & \cellcolor[HTML]{F3F2C5}\textbf{0.665} & \cellcolor[HTML]{F3F2C5}13 & \cellcolor[HTML]{F3F2C5}0.665 & \cellcolor[HTML]{F3F2C5}23 & \cellcolor[HTML]{F3F2C5}0.630 & \cellcolor[HTML]{F3F2C5}4 & \cellcolor[HTML]{F3F2C5}0.630 & \cellcolor[HTML]{F3F2C5}4 \\ \cline{2-12} 
\multirow{-5}{*}{\textbf{DD}} & \cellcolor[HTML]{F3F2C5}$\textbf{w/ Gen.}_{3}$ = $|R|*3$ & \cellcolor[HTML]{F3F2C5}0.630 & \cellcolor[HTML]{F3F2C5}14 & \cellcolor[HTML]{F3F2C5}\textbf{0.665} & \cellcolor[HTML]{F3F2C5}15 & \cellcolor[HTML]{F3F2C5}\textbf{0.685} & \cellcolor[HTML]{F3F2C5}12 & \cellcolor[HTML]{F3F2C5}0.630 & \cellcolor[HTML]{F3F2C5}5 & \cellcolor[HTML]{F3F2C5}0.630 & \cellcolor[HTML]{F3F2C5}5 \\ \hline

 & raw-data & 0.576 & 7 & 0.718 & 40 & 0.707 & 11 & 0.534 & 8 & 0.650 & 5 \\ \cline{2-12} 
 & w/ Real & 0.572 & 10 & 0.736 & 28 & \textbf{0.736} & 10 & \textbf{0.693} & 13 & 0.656 & 16 \\ \cline{2-12} 
 & \cellcolor[HTML]{F3F2C5}$\textbf{w/ Gen.}_{1}$& \cellcolor[HTML]{F3F2C5}\textbf{0.578} & \cellcolor[HTML]{F3F2C5}8 & \cellcolor[HTML]{F3F2C5}0.705 & \cellcolor[HTML]{F3F2C5}9 & \cellcolor[HTML]{F3F2C5}0.718 & \cellcolor[HTML]{F3F2C5}11 & \cellcolor[HTML]{F3F2C5}0.534 & \cellcolor[HTML]{F3F2C5}6 & \cellcolor[HTML]{F3F2C5}0.666 & \cellcolor[HTML]{F3F2C5}14 \\ \cline{2-12} 
 & \cellcolor[HTML]{F3F2C5}$\textbf{w/ Gen.}_{2}$ & \cellcolor[HTML]{F3F2C5}0.572 & \cellcolor[HTML]{F3F2C5}14 & \cellcolor[HTML]{F3F2C5}0.730 & \cellcolor[HTML]{F3F2C5}10 & \cellcolor[HTML]{F3F2C5}0.734 & \cellcolor[HTML]{F3F2C5}11 & \cellcolor[HTML]{F3F2C5}0.650 & \cellcolor[HTML]{F3F2C5}24 & \cellcolor[HTML]{F3F2C5}\textbf{0.701} & \cellcolor[HTML]{F3F2C5}19 \\ \cline{2-12} 
\multirow{-5}{*}{\textbf{COLLAB}} & \cellcolor[HTML]{F3F2C5}$\textbf{w/ Gen.}_{3}$ & \cellcolor[HTML]{F3F2C5}0.569 & \cellcolor[HTML]{F3F2C5}13 & \cellcolor[HTML]{D3D050}\textbf{0.738} & \cellcolor[HTML]{F3F2C5}33 & \cellcolor[HTML]{F3F2C5}0.716 & \cellcolor[HTML]{F3F2C5}8 & \cellcolor[HTML]{F3F2C5}0.631 & \cellcolor[HTML]{F3F2C5}13 & \cellcolor[HTML]{F3F2C5}0.627 & \cellcolor[HTML]{F3F2C5}23 \\ \hline

 & raw-data & 0.576 & 16 & 0.681 & 20 & 0.701 & 39 & 0.689 & 30 & 0.682 & 45 \\ \cline{2-12} 
 & w/ Real & 0.576 & 7 & 0.694 & 31 & 0.700 & 30 & 0.684 & 13 & \multicolumn{2}{c}{OOM} \\ \cline{2-12} 
 & \cellcolor[HTML]{F3F2C5}$\textbf{w/ Gen.}_{1}$ & \cellcolor[HTML]{F3F2C5}0.578 & \cellcolor[HTML]{F3F2C5}10 & \cellcolor[HTML]{F3F2C5}\textbf{0.700} & \cellcolor[HTML]{F3F2C5}24 & \cellcolor[HTML]{D3D050}\textbf{0.702} & \cellcolor[HTML]{F3F2C5}23 & \cellcolor[HTML]{F3F2C5}0.695 & \cellcolor[HTML]{F3F2C5}21 & \multicolumn{2}{c}{\cellcolor[HTML]{F3F2C5}OOM} \\ \cline{2-12} 
 & \cellcolor[HTML]{F3F2C5}$\textbf{w/ Gen.}_{2}$ & \cellcolor[HTML]{F3F2C5}\textbf{0.579} & \cellcolor[HTML]{F3F2C5}12 & \cellcolor[HTML]{F3F2C5}0.699 & \cellcolor[HTML]{F3F2C5}12 & \cellcolor[HTML]{D3D050}\textbf{0.702} & \cellcolor[HTML]{F3F2C5}28 & \cellcolor[HTML]{F3F2C5}0.696 & \cellcolor[HTML]{F3F2C5}19 & \multicolumn{2}{c}{\cellcolor[HTML]{F3F2C5}OOM} \\ \cline{2-12} 
\multirow{-5}{*}{\textbf{TWITCH EGOS}} & \cellcolor[HTML]{F3F2C5}$\textbf{w/ Gen.}_{3}$ & \cellcolor[HTML]{F3F2C5}0.575 & \cellcolor[HTML]{F3F2C5}9 & \cellcolor[HTML]{F3F2C5}0.699 & \cellcolor[HTML]{F3F2C5}13 & \cellcolor[HTML]{F3F2C5}0.690 & \cellcolor[HTML]{F3F2C5}42 & \cellcolor[HTML]{F3F2C5}\textbf{0.700} & \cellcolor[HTML]{F3F2C5}35 & \multicolumn{2}{c}{\cellcolor[HTML]{F3F2C5}OOM} \\ \hline

 & raw-data & 0.597 & 20 & 0.703 & 15 & \textbf{0.740} & 28 & 0.551 & 4 & 0.701 & 28 \\ \cline{2-12} 
 & w/ Real & 0.590 & 14 & 0.726 & 16 & 0.719 & 12 & 0.551 & 4 & \textbf{0.719} & 44 \\ \cline{2-12} 
 & \cellcolor[HTML]{F3F2C5}$\textbf{w/ Gen.}_{1}$ & \cellcolor[HTML]{F3F2C5}0.593 & \cellcolor[HTML]{F3F2C5}29 & \cellcolor[HTML]{F3F2C5}0.698 & \cellcolor[HTML]{F3F2C5}10 & \cellcolor[HTML]{F3F2C5}0.726 & \cellcolor[HTML]{F3F2C5}22 & \cellcolor[HTML]{F3F2C5}0.551 & \cellcolor[HTML]{F3F2C5}4 & \cellcolor[HTML]{F3F2C5}0.717 & \cellcolor[HTML]{F3F2C5}30 \\ \cline{2-12} 
 & \cellcolor[HTML]{F3F2C5}$\textbf{w/ Gen.}_{2}$& \cellcolor[HTML]{F3F2C5}\textbf{0.609} & \cellcolor[HTML]{F3F2C5}18 & \cellcolor[HTML]{F3F2C5}0.708 & \cellcolor[HTML]{F3F2C5}11 & \cellcolor[HTML]{F3F2C5}0.728 & \cellcolor[HTML]{F3F2C5}14 & \cellcolor[HTML]{F3F2C5}0.551 & \cellcolor[HTML]{F3F2C5}4 & \cellcolor[HTML]{F3F2C5}0.701 & \cellcolor[HTML]{F3F2C5}32 \\ \cline{2-12} 
\multirow{-5}{*}{\textbf{MUTAGENICITY}} & \cellcolor[HTML]{F3F2C5}$\textbf{w/ Gen.}_{3}$ & \cellcolor[HTML]{F3F2C5}0.586 & \cellcolor[HTML]{F3F2C5}12 & \cellcolor[HTML]{D3D050}\textbf{0.744} & \cellcolor[HTML]{F3F2C5}32 & \cellcolor[HTML]{F3F2C5}0.712 & \cellcolor[HTML]{F3F2C5}31 & \cellcolor[HTML]{F3F2C5}0.551 & \cellcolor[HTML]{F3F2C5}4 & \cellcolor[HTML]{F3F2C5}0.685 & \cellcolor[HTML]{F3F2C5}22 \\ \hline

 & raw-data & 0.141 & 9 & 0.166 & 7 & 0.191 & 46 & 0.166 & 13 & 0.166 & 10 \\ \cline{2-12} 
 & w/ Real & 0.166 & 8 & 0.158 & 15 & 0.166 & 96 & 0.166 & 8 & 0.166 & 17 \\ \cline{2-12} 
 & \cellcolor[HTML]{F3F2C5}$\textbf{w/ Gen.}_{1}$ & \cellcolor[HTML]{F3F2C5}\textbf{0.233} & \cellcolor[HTML]{F3F2C5}14 & \cellcolor[HTML]{F3F2C5}0.200 & \cellcolor[HTML]{F3F2C5}13 & \cellcolor[HTML]{D3D050}\textbf{0.266} & \cellcolor[HTML]{F3F2C5}41 & \cellcolor[HTML]{F3F2C5}0.166 & \cellcolor[HTML]{F3F2C5}10 & \cellcolor[HTML]{F3F2C5}\textbf{0.175} & \cellcolor[HTML]{F3F2C5}15 \\ \cline{2-12} 
 & \cellcolor[HTML]{F3F2C5}$\textbf{w/ Gen.}_{2}$ & \cellcolor[HTML]{F3F2C5}0.191 & \cellcolor[HTML]{F3F2C5}7 & \cellcolor[HTML]{F3F2C5}\textbf{0.233} & \cellcolor[HTML]{F3F2C5}27 & \cellcolor[HTML]{D3D050}\textbf{0.266} & \cellcolor[HTML]{F3F2C5}35 & \cellcolor[HTML]{F3F2C5}0.166 & \cellcolor[HTML]{F3F2C5}12 & \cellcolor[HTML]{F3F2C5}\textbf{0.175} & \cellcolor[HTML]{F3F2C5}15 \\ \cline{2-12} 
\multirow{-5}{*}{\textbf{ENZYMES}} & \cellcolor[HTML]{F3F2C5}$\textbf{w/ Gen.}_{3}$& \cellcolor[HTML]{F3F2C5}0.175 & \cellcolor[HTML]{F3F2C5}11 & \cellcolor[HTML]{F3F2C5}0.208 & \cellcolor[HTML]{F3F2C5}18 & \cellcolor[HTML]{F3F2C5}0.158 & \cellcolor[HTML]{F3F2C5}22 & \cellcolor[HTML]{F3F2C5}0.166 & \cellcolor[HTML]{F3F2C5}11 & \cellcolor[HTML]{F3F2C5}0.166 & \cellcolor[HTML]{F3F2C5}15 \\ \hline

 & raw-data & 0.666 & 22 & 0.944 & 25 & 0.944 & 26 & 0.666 & 10 & 0.666 & 11 \\ \cline{2-12} 
 & w/ Real & 0.666 & 32 & 0.833 & 54 & 0.833 & 15 & 0.666 & 11 & 0.666 & 10 \\ \cline{2-12} 
 & \cellcolor[HTML]{F3F2C5}$\textbf{w/ Gen.}_{1}$ & \cellcolor[HTML]{F3F2C5}0.666 & \cellcolor[HTML]{F3F2C5}12 & \cellcolor[HTML]{D3D050}\textbf{0.944} & \cellcolor[HTML]{F3F2C5}42 & \cellcolor[HTML]{F3F2C5}0.833 & \cellcolor[HTML]{F3F2C5}14 & \cellcolor[HTML]{F3F2C5}0.666 & \cellcolor[HTML]{F3F2C5}11 & \cellcolor[HTML]{F3F2C5}0.666 & \cellcolor[HTML]{F3F2C5}11 \\ \cline{2-12} 
 & \cellcolor[HTML]{F3F2C5}$\textbf{w/ Gen.}_{2}$ & \cellcolor[HTML]{F3F2C5}0.666 & \cellcolor[HTML]{F3F2C5}11 & \cellcolor[HTML]{D3D050}\textbf{0.944} & \cellcolor[HTML]{F3F2C5}51 & \cellcolor[HTML]{D3D050}\textbf{0.944} & \cellcolor[HTML]{F3F2C5}24 & \cellcolor[HTML]{F3F2C5}0.666 & \cellcolor[HTML]{F3F2C5}11 & \cellcolor[HTML]{F3F2C5}0.666 & \cellcolor[HTML]{F3F2C5}10 \\ \cline{2-12} 
\multirow{-5}{*}{\textbf{MUTAG}} & \cellcolor[HTML]{F3F2C5}$\textbf{w/ Gen.}_{3}$ & \cellcolor[HTML]{F3F2C5}0.666 & \cellcolor[HTML]{F3F2C5}40 & \cellcolor[HTML]{F3F2C5}0.833 & \cellcolor[HTML]{F3F2C5}25 & \cellcolor[HTML]{D3D050}\textbf{0.944} & \cellcolor[HTML]{F3F2C5}36 & \cellcolor[HTML]{F3F2C5}0.666 & \cellcolor[HTML]{F3F2C5}13 & \cellcolor[HTML]{F3F2C5}0.666 & \cellcolor[HTML]{F3F2C5}11 \\ \hline
\end{tabular}
}
\end{table*}

To investigate the impact of generated graphs on the graph classification task, we partitioned each dataset into three, ensuring the class distributions remained consistent: raw-data (80\%), sub-real data (10\%), and test data (10\%). The raw-data serves as the baseline for comparison, representing the initial data available to researchers. In comparison, the sub-real data ($R$) imitates the supplementary data that researchers might acquire through additional time and resource investment in real-world scenarios. Lastly, the test data is designated to evaluate the performance of graph classification. Created with real-world application in mind, this strategic data partitioning enables the analysis of how generated graphs affect the accuracy of graph classification models.

Leveraging the initial raw-data available to researchers, we generated datasets comparable in size to the $R$, ensuring class distributions remained consistent across all generated sets as in all other sets. This approach allowed us to assess the impact of the real and generated data of the same size on graph classification performance. We further extended these experiments by generating data sets twice and three times the size of the $R$, to examine how change in the volume of generated graphs affects the model performance. Given the diversity in graph sizes within our datasets (see. varying sizes of avg. nodes, avg. edges in Table \ref{tab:data}), we employed GRAN, known for its adaptability to large graphs, as our preliminary study indicated that GraphRNN encountered Out of Memory (OOM) errors with bigger graphs. Table \ref{tab:res} presents the prediction results of graph classifier models from various backgrounds for raw-data, with $R$ added to the raw-data (w/ Real), and with the aforementioned generated data sets added to the raw-data namely ${w/ Gen.}_{1}$, ${w/ Gen.}_{2}$, ${w/ Gen.}_{3}$. According to the results in Table \ref{tab:res}, the most accurate predictions for each dataset, as highlighted in the table, were achieved by incorporating the generated graphs and mostly with ${w/ Gen.}_{2}$.

In the second part of the experiments, we generated one thousand twenty-four graphs from each class in the datasets and drastically increased the proportion of the generated data size. Here, we aim to explore the feasibility of obtaining a more balanced dataset with a large number of samples with the proposed graph data generation method, avoiding the imbalanced data and data scarcity problems that affect the prediction performance of many learning algorithms. The results of these experiments, which we obtained by data generation with GraphRNN and GRAN on datasets MUTAGENICITY, ENZYMES, and MUTAG, which consist of small graphs and contain a small number of samples, are presented in Table \ref{tab:res2}. According to the results in Table \ref{tab:res2}, the most accurate predictions for each dataset, as highlighted in the table, were achieved with ${w/ Gen. (GraphRNN)}$.

\begin{table*}[ht]
\caption{Graph Classification Results - II}
\renewcommand{\arraystretch}{1.2}
\label{tab:res2}
\centering
\resizebox{.99\linewidth}{!}{
\begin{tabular}{clcccccccccc}
\hline
\textbf{} &  & \multicolumn{2}{c}{\textbf{GraphSAGE}} & \multicolumn{2}{c}{\textbf{GIN0}} & \multicolumn{2}{c}{\textbf{GINWithJK}} & \multicolumn{2}{c}{\textbf{GCNWithJK}} & \multicolumn{2}{c}{\textbf{EdgePool}} \\ \hline
\textbf{} &  & \textbf{Acc.} & \textbf{Epoch} & \textbf{Acc.} & \textbf{Epoch} & \textbf{Acc.} & \textbf{Epoch} & \textbf{Acc.} & \textbf{Epoch} & \textbf{Acc.} & \textbf{Epoch} \\ \hline
 & raw-data & 0.597 & 20 & 0.703 & 15 & 0.740 & 28 & 0.551 & 4 & 0.701 & 28 \\ \cline{2-12} 
 & w/ Real & 0.590 & 14 & \textbf{0.726} & 16 & 0.719 & 12 & 0.551 & 4 & \textbf{0.719} & 44 \\ \cline{2-12} 
 & \cellcolor[HTML]{F3F2C5}$\textbf{w/ Gen. (GraphRNN)}$ & \cellcolor[HTML]{F3F2C5}\textbf{0,611} & \cellcolor[HTML]{F3F2C5}20 & \cellcolor[HTML]{F3F2C5}0,710 & \cellcolor[HTML]{F3F2C5}14 & \cellcolor[HTML]{D3D050}\textbf{0,744} & \cellcolor[HTML]{F3F2C5}23 & \cellcolor[HTML]{F3F2C5}0,551 & \cellcolor[HTML]{F3F2C5}6 & \cellcolor[HTML]{F3F2C5}0,694 & \cellcolor[HTML]{F3F2C5}35 \\ \cline{2-12} 
\multirow{-4}{*}{\textbf{MUTAGENICITY}} & \cellcolor[HTML]{F3F2C5}$\textbf{w/ Gen. (GRAN)}$ & \cellcolor[HTML]{F3F2C5}0,551 & \cellcolor[HTML]{F3F2C5}7 & \cellcolor[HTML]{F3F2C5}0,689 & \cellcolor[HTML]{F3F2C5}21 & \cellcolor[HTML]{F3F2C5}0.728 & \cellcolor[HTML]{F3F2C5}23 & \cellcolor[HTML]{F3F2C5}0,551 & \cellcolor[HTML]{F3F2C5}5 & \cellcolor[HTML]{F3F2C5}0,551 & \cellcolor[HTML]{F3F2C5}6 \\ \hline
 & raw-data & 0.141 & 9 & 0.166 & 7 & 0.191 & 46 & 0.166 & 13 & 0.166 & 10 \\ \cline{2-12} 
 & w/ Real & 0.166 & 8 & 0.158 & 15 & 0.166 & 96 & 0.166 & 8 & 0.166 & 17 \\ \cline{2-12} 
 & \cellcolor[HTML]{F3F2C5}$\textbf{w/ Gen. (GraphRNN)}$ & \cellcolor[HTML]{F3F2C5}0,166 & \cellcolor[HTML]{F3F2C5}12 & \cellcolor[HTML]{D3D050}\textbf{0,241} & \cellcolor[HTML]{F3F2C5}49 & \cellcolor[HTML]{D3D050}\textbf{0,241} & \cellcolor[HTML]{F3F2C5}65 & \cellcolor[HTML]{F3F2C5}0,166 & \cellcolor[HTML]{F3F2C5}14 & \cellcolor[HTML]{F3F2C5}0,166 & \cellcolor[HTML]{F3F2C5}15 \\ \cline{2-12} 
\multirow{-4}{*}{\textbf{ENZYMES}} & \cellcolor[HTML]{F3F2C5}$\textbf{w/ Gen. (GRAN)}$ & \cellcolor[HTML]{F3F2C5}\textbf{0,183} & \cellcolor[HTML]{F3F2C5}7 & \cellcolor[HTML]{F3F2C5}0,158 & \cellcolor[HTML]{F3F2C5}31 & \cellcolor[HTML]{F3F2C5}0,125 & \cellcolor[HTML]{F3F2C5}75 & \cellcolor[HTML]{F3F2C5}0,166 & \cellcolor[HTML]{F3F2C5}22 & \cellcolor[HTML]{F3F2C5}\textbf{0,175} & \cellcolor[HTML]{F3F2C5}30 \\ \hline
 & raw-data & 0.666 & 22 & \textbf{0.944} & 25 & 0.944 & 26 & 0.666 & 10 & 0.666 & 11 \\ \cline{2-12} 
 & w/ Real & 0.666 & 32 & 0.833 & 54 & 0.944 & 15 & 0.666 & 11 & 0.666 & 10 \\ \cline{2-12} 
 & \cellcolor[HTML]{F3F2C5}$\textbf{w/ Gen. (GraphRNN)}$ & \cellcolor[HTML]{F3F2C5}0,388 & \cellcolor[HTML]{F3F2C5}70 & \cellcolor[HTML]{F3F2C5}0,888 & \cellcolor[HTML]{F3F2C5}85 & \cellcolor[HTML]{D3D050}\textbf{0,944} & \cellcolor[HTML]{F3F2C5}79 & \cellcolor[HTML]{F3F2C5}0,666 & \cellcolor[HTML]{F3F2C5}14 & \cellcolor[HTML]{F3F2C5}\textbf{0,777} & \cellcolor[HTML]{F3F2C5}52 \\ \cline{2-12} 
\multirow{-4}{*}{\textbf{MUTAG}} & \cellcolor[HTML]{F3F2C5}$\textbf{w/ Gen. (GRAN)}$ & \cellcolor[HTML]{D3D050}\textbf{0,944} & \cellcolor[HTML]{F3F2C5}27 & \cellcolor[HTML]{F3F2C5}0,666 & \cellcolor[HTML]{F3F2C5}30 & \cellcolor[HTML]{F3F2C5}0,777 & \cellcolor[HTML]{F3F2C5}60 & \cellcolor[HTML]{F3F2C5}0,666 & \cellcolor[HTML]{F3F2C5}14 & \cellcolor[HTML]{F3F2C5}0,777 & \cellcolor[HTML]{F3F2C5}21 \\ \hline
\end{tabular}
}
\end{table*}

\begin{figure*}[ht]
    \centering
    \includegraphics[width=.99\linewidth]{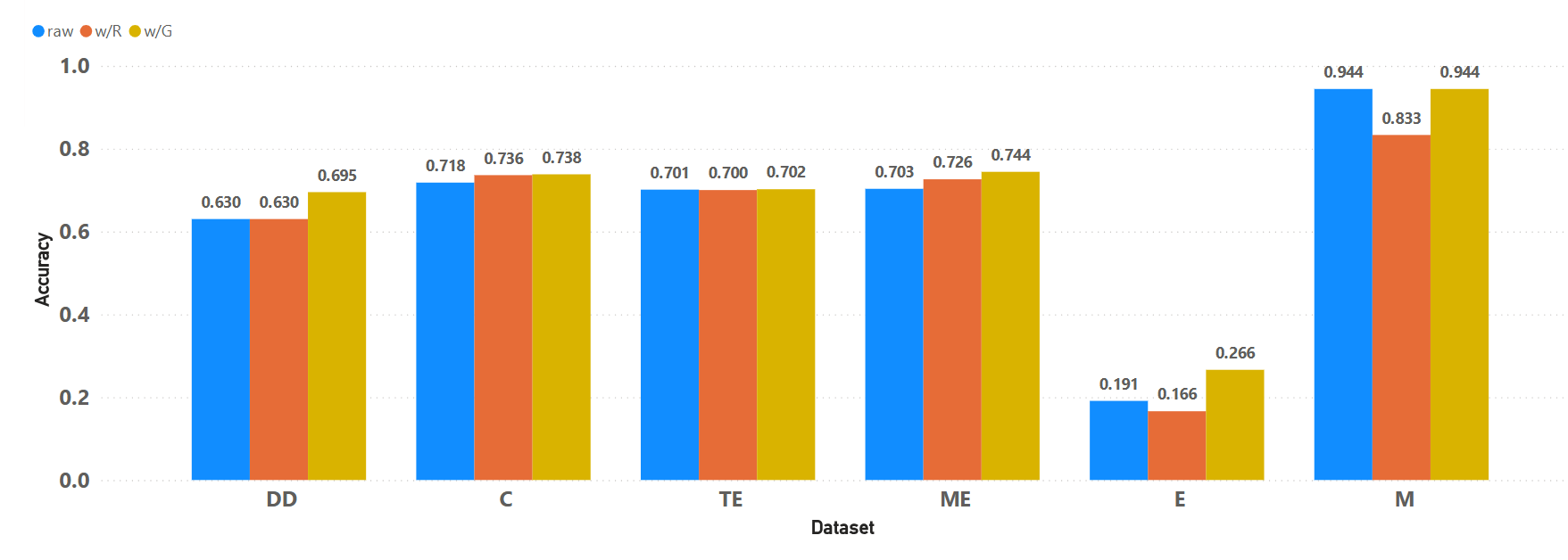}
    \caption{Proposed Graph Classification with Graph Size-aware Data Augmentation Framework Results Summary}
    \label{fig:res}
\end{figure*}

The overall summary of the results we obtained with the proposed textit{Graph classification with graph size-aware data augmentation} framework is presented in Fig. \ref{fig:res}. The results here reflect the raw-data, w/R and w/G accuracy values of the most accurate graph classifier for the relevant dataset (DD, COLLAB (C), TWITCH EGOS (TE), MUTAGENICITY (ME), ENZYMES (E), MUTAG (M)). Mirroring the observations of Touat et al. \cite{touat2023gran}, our study reveals that particularly when working with medium to large-sized graphs, the graphs produced by GRAN are more analogous to the original graphs, however, GraphRNN has scalability problems not applicable to large graphs. However, while working with smaller graphs GRAN tends to overfit and its generation quality drops, hence GrapRNN is better.

\section{Conclusion and Future Work}  \label{sec:6}

To conclude that, our study demonstrates the substantial impact of synthetic graph data on the performance of graph classification tasks across diverse datasets. The proposed \textit{Graph classification with graph size-aware data augmentation} framework offers a flexible graph data generation process that is applicable for small, medium, and large-sized graphs. Moreover, the experiments involving a significant increase in the proportion of generated data further highlighted the potential of our graph generation framework to mitigate issues of data imbalance and scarcity, especially in smaller datasets. This balance was crucial for achieving higher prediction accuracy, as evidenced by the superior performance of models utilizing balanced, generated datasets. Overall, our findings underscore the effectiveness of AI-driven data generation in enhancing graph classification tasks, paving the way for more accurate and reliable machine learning models in diverse applications. 

For future work, we plan to investigate the reasons for the differences in the performance of the generated data with explainable AI methods and also to work on generating synthetic graphs for dynamic graphs.

\begin{credits}
\subsubsection{\ackname} This research is supported by the Scientific and Technological Research Council of Turkey (TUBITAK) 1515 Frontier R\&D Laboratories Support Program (project number 5239903) and the ITU Scientific Research Projects Fund under grant number YESAP-2024-45920.

\subsubsection{\discintname}
There are no relevant financial or non-financial competing interests to report.
\end{credits}

\bibliographystyle{unsrt}
\bibliography{bibliography.bib}
\end{document}